\pdfoutput=1
\documentclass[11pt]{article}

\usepackage[a4paper,margin=2.5cm]{geometry}
\usepackage[T1]{fontenc}
\usepackage[utf8]{inputenc}
\usepackage{graphicx}
\usepackage{amsmath,amssymb,amsfonts}
\usepackage{bm}
\usepackage{booktabs}
\usepackage{microtype}
\usepackage{balance}
\usepackage{tikz}
\usepackage[numbers,square]{natbib}

\usepackage[
  colorlinks=true,
  urlcolor=blue,
  linkcolor=blue,
  citecolor=blue
]{hyperref}

\definecolor{gaussblue}{RGB}{95,140,190}
\definecolor{rkhsorange}{RGB}{225,155,80}
\definecolor{compatgreen}{RGB}{105,165,115}

\newcommand{\HK}{\mathcal{H}_K}
\newcommand{\ellone}{\ell_1}
\newcommand{\elltwo}{\ell_2}

\newtheorem{definition}{Definition}

\newtheorem{theorem}{Theorem}
\newtheorem{proposition}{Proposition}
\newtheorem{corollary}{Corollary}

\newenvironment{proof}[1][Proof]{%
  \par\noindent\textit{#1.}\ }%
  {\hfill$\square$\par}

\allowdisplaybreaks

\title{Stability-compatible kernels: beyond Gaussianity\\ for impulse response estimation}
\author{Gianluigi Pillonetto\thanks{Corresponding author.} \quad Mauro Bisiacco\\[0.6ex]
\small Department of Information Engineering, University of Padova\\
\small Via Gradenigo 6/B, 35131 Padova, Italy\\
\small \texttt{giapi@dei.unipd.it}, \texttt{bisiacco@dei.unipd.it}}
\date{}

\begin{document}

\maketitle

\begin{abstract}
Kernel-based regularization for linear system identification uses positive semidefinite kernels both to define reproducing kernel Hilbert spaces (RKHSs) and as covariance models for stochastic impulse responses. Sharp kernel conditions are known for RKHS stability, while almost-sure BIBO stability under Gaussian priors requires an even more restrictive covariance condition. These restrictions, however, need not persist beyond Gaussianity, where the covariance no longer determines the full probability law. We introduce the notion of a stability-compatible kernel, namely a covariance kernel admitting at least one stochastic realization with almost-surely absolutely summable sample paths. We show that stability-compatibility holds if and only if the stable elements are dense in the associated RKHS, a condition strictly weaker than RKHS stability, and derive practical sufficient conditions, including a spectral criterion for stationary kernels.
We then construct non-Gaussian priors that preserve a prescribed covariance while enforcing almost-sure stability. In particular, this yields a stable prior based on the stationary radial-basis-function (RBF) kernel, separating covariance structure from the mechanism enforcing stability. Numerical experiments show substantial estimation gains for several important classes of impulse responses, with particularly pronounced advantages when relevant dynamics occur at late lags or in separated temporal regions.
\end{abstract}

\medskip

\section{Introduction}
\label{sec:introduction}

Kernel-based regularization has become a standard nonparametric tool for linear system identification
\citep{PillonettoDeNicolao2010,RegularizedSI2022,NatRev2026}.
A positive semidefinite kernel plays two closely related roles in this framework.
Deterministically, it defines a reproducing kernel Hilbert space (RKHS) and hence the geometry
of the regularization problem \citep{Aronszajn1950,CuckerSmale2002}.
Probabilistically, the same kernel can be interpreted as the covariance of a stochastic prior
on the unknown impulse response, most commonly a Gaussian process
\citep{ChenOhlssonLjung2012,RasmussenWilliams2006}.
This dual deterministic--probabilistic interpretation is one of the main attractions of kernel
methods \citep{ScholkopfSmola2002,BerlinetThomasAgnan2004,NatCom2026}, but it also raises a subtle issue
when structural properties of dynamical systems, such as stability, are considered.

For discrete-time linear systems, BIBO stability requires the impulse response to be absolutely summable. In the deterministic RKHS viewpoint, this has traditionally led to the notion of a stable kernel, requiring all elements of the associated RKHS to be stable. The corresponding theory is by now well developed and includes several characterizations and kernel-design principles specifically tailored to system identification \citep{PillonettoDeNicolao2010,ChenOhlssonLjung2012,BisiaccoPillonetto2020}. From a probabilistic perspective, however, the requirement is imposed on the sample paths of the stochastic process. For Gaussian processes, almost-sure BIBO stability is more restrictive than RKHS stability and is completely determined by the covariance through the diagonal square-root summability condition identified in \citep{KhosraviSmith2023}.

This distinction becomes important because Gaussianity ties covariance and sample-path behavior together. Once the mean and covariance are fixed, the Gaussian law is fixed as well, so stability cannot be imposed independently of the second-order structure. Beyond Gaussianity, this restriction disappears: distinct stochastic processes may have exactly the same covariance while possessing very different higher-order structure and sample-path properties. This opens the possibility of using the covariance to describe dependence or smoothness while enforcing stability through the full prior distribution rather than through a prescribed covariance decay. A classical result of \citep{LukicBeder2001} already implies that every RKHS-stable kernel admits a generally non-Gaussian realization whose sample paths belong to the RKHS and are therefore stable. This leaves open the more general situation in which the RKHS itself contains unstable elements.

In view of this, the first question addressed in this paper is structural: to characterize the covariance kernels that admit at least one stochastic realization with stable sample paths almost surely. We call such kernels \emph{stability-compatible}. Our main result shows that stability-compatibility holds precisely when the stable elements of the associated RKHS are dense in the RKHS norm. This condition is strictly weaker than conventional RKHS stability and therefore identifies a strictly larger class of covariance structures supporting stable stochastic impulse responses, including kernels whose associated Gaussian processes are unstable and whose RKHSs contain unstable elements. We complement this characterization with practical sufficient conditions capturing different mechanisms leading to stability-compatibility. These include general kernel-level criteria and, for stationary kernels, a spectral condition based on absolute continuity of the spectral measure together with the classical Szeg\H{o} condition  \citep{Bingham2012}.  We also develop explicit non-Gaussian priors that preserve a prescribed covariance while enforcing almost-sure stability.

The second question is statistical: whether the additional freedom offered by non-Gaussian priors can translate into improved impulse-response estimation. We specialize our construction to the stationary radial-basis-function (RBF) covariance and obtain a stable non-Gaussian RBF prior. In this model, the covariance describes local dependence while stability is enforced separately through the higher-order structure of the prior. This differs from classical kernels developed for stable system identification, such as the tuned/correlated (TC) kernel, where stability is encoded directly through covariance decay \citep{PillonettoDeNicolao2010,ChenOhlssonLjung2012}. Our numerical study shows that this different modeling principle can provide clear estimation benefits across a range of stable impulse responses, with the largest gains arising when significant dynamics are located at late lags or distributed over separated temporal regions.

The paper is organized as follows.
Section~\ref{sec:estimation} introduces the impulse-response estimation problem and discusses the role of the prior, using the TC and RBF covariances to motivate the separation between second-order structure and the mechanism enforcing stability.
Section~\ref{sec:preliminaries} formalizes the three notions of Gaussian stability, RKHS stability, and stability-compatibility.
Section~\ref{sec:mainresult} establishes the necessary and sufficient characterization of stability-compatible kernels. Section~\ref{sec:sufficient} derives practical sufficient conditions, illustrates them through representative examples including RBF and ARMA covariance models, and develops a spectral criterion for stationary kernels.
Section~\ref{sec:hierarchy} establishes the strict hierarchy among the three notions of stability.
Section~\ref{sec:twoscale} develops a constructive family of stable non-Gaussian priors with prescribed covariance and specializes it to the stable RBF prior.
Section~\ref{sec:numexp} evaluates the resulting estimator numerically and compares it with standard Gaussian kernel methods, with particular emphasis on the TC prior. Conclusions then end the paper.
%Finally, Section~\ref{sec:discussion} discusses the implications of these results for kernel design and probabilistic system identification.

\section{Impulse response estimation and the role of the prior}
\label{sec:estimation}

We first place the problem studied in this paper in the standard setting of nonparametric linear system identification. Consider a causal, discrete-time, linear time-invariant system
\begin{equation}
    y_t=\sum_{k=1}^{\infty} g_k u_{t-k}+e_t,
    \label{eq:iir-model}
\end{equation}
where $u_t$ is the known input, $y_t$ is the measured output, $e_t$ is measurement noise, and
\[
g=(g_1,g_2,\ldots)
\]
is the unknown impulse response. We are interested in IIR models, so no finite support assumption is imposed on $g$. For a BIBO-stable system, the impulse response satisfies $g\in\ell_1$.

Let
\[
K:\mathbb N\times\mathbb N\to\mathbb R
\]
be a positive semidefinite kernel, i.e.,
\[
\sum_{i,j=1}^{N} c_i c_j K(n_i,n_j)\geq 0
\]
for every $N\geq1$, every choice of indices $n_1,\ldots,n_N\in\mathbb N$, and every $c_1,\ldots,c_N\in\mathbb R$. We denote by $\mathcal H_K$ the reproducing kernel Hilbert space associated with $K$ \citep{Aronszajn1950}.

Given a finite collection of input-output measurements, a standard kernel-based estimate of $g$ is obtained by solving
\begin{equation}
    \widehat g
    =
    \arg\min_{g\in\mathcal H_K}
    \left\{
        \sum_{t=1}^{T}
        \Big(
            y_t-\sum_{k=1}^{\infty}g_k u_{t-k}
        \Big)^2
        +
        \lambda \|g\|_{\mathcal H_K}^2
    \right\},
    \label{eq:kernel-estimator}
\end{equation}
where $\|\cdot\|_{\mathcal H_K}$ is the RKHS norm and $\lambda>0$ controls the amount of regularization. Thus, the kernel determines the geometry of the regularizer and therefore which impulse responses are favored by the estimator.

The same kernel also admits a probabilistic interpretation. Since $K$ is positive semidefinite, it can be used as the covariance of a zero-mean stochastic process $g$, so that
\begin{equation}
    \mathbb E[g_i g_j]=K(i,j).
    \label{eq:prior-covariance}
\end{equation}
Assume that the prior is Gaussian and that the measurement noise is also Gaussian and independent of $g$. Then the regularized estimate associated with \eqref{eq:kernel-estimator} coincides, with the regularization parameter determined by the ratio between the noise variance and the prior scale, with the posterior mean
\begin{equation}
    \widehat g=\mathbb E[g\mid y].
    \label{eq:posterior-mean}
\end{equation}
It is therefore the minimum mean-square error estimator and, under these Gaussian assumptions, is linear in the observed outputs.

This probabilistic viewpoint is particularly natural for kernels developed specifically for stable system identification, including the tuned/correlated (TC), diagonal/correlated (DC), and stable-spline kernels \citep{PillonettoDeNicolao2010,ChenOhlssonLjung2012}. Apart from an overall scale parameter, the TC kernel is
\begin{equation}
    K_{\rm TC}(i,j)=\beta^{\max(i,j)},
    \qquad 0<\beta<1.
    \label{eq:tc-kernel}
\end{equation}
Since
\[
K_{\rm TC}(n,n)=\beta^n,
\]
the marginal variance $\operatorname{Var}(g_n)=\beta^n$ decreases exponentially with the lag, while the off-diagonal entries encode temporal dependence among the impulse-response coefficients. Stable-spline and DC kernels follow the same broad modeling principle, combining stability with additional regularity or correlation structure through the covariance. This line of research has subsequently led to more systematic kernel-design frameworks, maximum-entropy characterizations, frequency-domain interpretations, and higher-order generalizations of classical kernels \citep{CarliChenLjung2017,Chen2018,ZorziChiuso2018,Zorzi2024}. Related extensions have also developed Gaussian-process kernel methods directly in the frequency domain for transfer-function estimation \citep{LataireChen2016}.

These observations motivate considering non-Gaussian priors that encode structural information on stable impulse responses beyond what is represented by the covariance kernel. Such constructions have received comparatively little attention. They are potentially important because the best linear estimator depends only on second-order statistics and therefore cannot distinguish between different priors sharing the same covariance. By contrast, the posterior mean depends on the full prior law and, for a non-Gaussian prior, is generally a nonlinear function of the observed outputs.

This distinction becomes particularly transparent for the radial-basis-function (RBF), or squared-exponential, kernel
\begin{equation}
    K_{\rm RBF}(i,j)
    =
    \exp\!\left(
        -\alpha (i-j)^2 %}{2\ell^2}
    \right), \quad \alpha>0,
    \label{eq:rbf-kernel-intro}
\end{equation}
one of the most widely used kernels in machine learning. Unlike TC, DC, and stable-spline kernels, the RBF kernel does not encode decay through its diagonal, since
\[
K_{\rm RBF}(n,n)=1
\]
for every $n$. Consequently, any zero-mean stochastic process with this covariance satisfies
\[
\operatorname{Var}(g_n)=1
\qquad\text{for every }n.
\]
If the process is Gaussian, this covariance structure is incompatible with almost-sure stability \citep{KhosraviSmith2023}. Thus, under the conventional Gaussian-process interpretation, the RBF kernel cannot define a prior supported on stable impulse responses.

From a modeling viewpoint, however, a nondecaying covariance diagonal can be desirable. Stability requires the impulse response to be absolutely summable, but does not prescribe a particular decay envelope or require its energy to be concentrated at early lags. Stable impulse responses may exhibit delayed components, resonances, multiple time scales, or localized activity at relatively large lags. A covariance whose diagonal decreases with the lag expresses the stronger prior belief that late coefficients should also become progressively smaller at the second-order level. When such information is not available, it may be preferable to use the covariance to describe dependence among neighboring coefficients while enforcing stability through other features of the prior law.

A central result of this paper shows that the incompatibility above is a consequence of Gaussianity rather than of the RBF covariance itself. The same RBF covariance admits a non-Gaussian stochastic realization whose sample paths belong to $\ell_1$ almost surely. More generally, this suggests a modeling principle in which second-order dependence and almost-sure stability are controlled separately: the covariance describes local dependence, while stability is encoded through higher-order features of the prior distribution.

This motivates the general problem studied in the next sections: characterizing the covariance kernels that admit at least one stochastic realization with stable sample paths almost surely. We begin by distinguishing three notions of stability associated with a kernel.

\section{Three notions of stability}
\label{sec:preliminaries}

We now formalize three distinct ways in which the kernel $K$ introduced above can be related to stable impulse responses.

As for every RKHS, the linear span of the kernel sections
\[
\{K(\cdot,n):n\in\mathbb N\}
\]
is dense in $\mathcal H_K$, 
i.e. its closure w.r.t. the RKHS norm satisfies 
\begin{equation}
\mathcal H_K=
\overline{\operatorname{span}\{K(\cdot,n):n\in\mathbb N\}}^{\,\mathcal H_K}
\label{eq:sectionspan}
\end{equation}
Since $\mathbb N$ is countable, $\mathcal H_K$ is separable.

On the probabilistic side, any zero-mean stochastic process $g=(g_n)_{n\geq1}$ with covariance $K$ satisfies
\begin{equation}
    \mathbb E[g_i g_j]=K(i,j),
    \qquad i,j\in\mathbb N.
    \label{eq:covdef}
\end{equation}
Throughout the paper, this only requires each component $g_n$ to have finite second moment.

Since BIBO stability is equivalent to $g\in\ell_1$, we distinguish the following three notions
describing how a kernel can be related to stable impulse responses.
The first one requires that  every impulse response belonging to the RKHS associated with $K$ is BIBO stable.

\begin{definition}[Stable RKHS kernel]
A kernel $K$ is said to be \emph{RKHS-stable}, or just
\emph{stable} in the system identification literature, if
\begin{equation}
    \mathcal H_K\subset\ell_1.
    \label{eq:stable-rkhs}
\end{equation}
\end{definition}

A classical necessary and sufficient condition for such RKHS stability is that the kernel operator
\[
(T_K u)_i=\sum_{j=1}^{\infty}K(i,j)u_j
\]
defines a bounded map from $\ell_\infty$ into $\ell_1$ \citep{CarmeliEtAl2006,PillonettoEtAl2014}.

\begin{definition}[Gaussian-stable kernel]
A kernel $K$ is said to be \emph{Gaussian-stable} if, for the zero-mean Gaussian process $g$ with covariance $K$,
\[
\mathbb{P}(g\in\ell_1)=1.
\]
\end{definition}

For a discrete-time zero-mean Gaussian process with covariance $K$,
Gaussian stability is equivalent to the diagonal square-root
integrability (DSRI) condition \citep{KhosraviSmith2023},
\begin{equation}
    \mathbb P(g\in\ell_1)=1
    \quad\Longleftrightarrow\quad
    \sum_{n=1}^{\infty}\sqrt{K(n,n)}<\infty.
    \label{eq:dsri}
\end{equation}
The DSRI condition implies RKHS stability in
\eqref{eq:stable-rkhs}, whereas the converse is false.

Finally, the following definition is new and will be the focus of this paper.

\begin{definition}[Stability-compatible kernel]
A kernel $K$ is \emph{stability-compatible} if there exists a zero-mean stochastic process \(g\) with covariance \(K\) such that
\begin{equation}
    \mathbb{P}(g\in\ell_1)=1.
    \label{eq:asstable}
\end{equation}
\end{definition}

Hence, stability-compatibility asks whether the covariance structure encoded by $K$ can arise from a stochastic model whose impulse responses are BIBO stable almost surely. In contrast with Gaussian stability, no particular distributional family is imposed: different stochastic processes may share the same covariance kernel while having very different sample-path properties.

We have seen that, for the first two notions, necessary and sufficient characterizations are already available. An analogous characterization of stability-compatibility remains to be established; this is the objective of the next section.

\section{Characterization of stability-compatible kernels}
\label{sec:mainresult}

We now seek a characterization of stability-compatibility directly in terms of the kernel $K$ and its associated RKHS. Recall that stability-compatibility only requires the existence of at least one stochastic process with covariance $K$ whose sample paths belong to $\ell_1$ almost surely; it does not require every element of $\mathcal H_K$ to be stable. The following result identifies exactly how much stability is required within the RKHS.

\begin{theorem}
\label{thm:main}
Let $K$ be a positive semidefinite kernel on $\mathbb{N}$. Then $K$ is stability-compatible if and only if
\begin{equation}
    \overline{\mathcal H_K\cap\ell_1}^{\,\mathcal H_K}
    =
    \mathcal H_K .
    \label{eq:density}
\end{equation}
\end{theorem}

\begin{proof}
We first prove necessity. Assume that there exists a zero-mean stochastic process
\[
g=(g_n)_{n\geq1}
\]
defined on a probability space $(\Omega,\mathcal F,\mathbb P)$, with covariance $K$, and such that
\[
g\in\ell_1
\qquad\text{almost surely}.
\]
Since
\[
\mathbb E[g_n^2]=K(n,n)<\infty,
\]
we have $g_n\in L^2(\Omega)$ for every $n\geq1$.

Let $\mathcal G$ denote the closed linear subspace of $L^2(\Omega)$ generated by the random variables $\{g_n\}_{n\geq1}$, namely
\[
\mathcal G
=
\overline{\operatorname{span}\{g_n:n\geq1\}}^{\,L^2(\Omega)}.
\]
The relevance of $\mathcal G$ follows from the fact that the covariance structure of the process reproduces the inner-product structure of the RKHS. Indeed,
\[
\langle g_i,g_j\rangle_{L^2(\Omega)}
=
\mathbb E[g_i g_j]
=
K(i,j),
\]
whereas, by the reproducing property,
\[
\langle K(\cdot,i),K(\cdot,j)\rangle_{\mathcal H_K}
=
K(i,j).
\]
Hence, following the standard RKHS--stochastic-process correspondence
\citep{Wahba:90}, the mapping
\[
K(\cdot,n)\longmapsto g_n
\]
preserves inner products on the linear span of the kernel sections. Since this span is dense in $\mathcal H_K$, the mapping extends uniquely by continuity to an isometric isomorphism
\[
\mathcal{I}:\mathcal H_K\longrightarrow\mathcal G
\]
satisfying
\[
\mathcal{I} K(\cdot,n)=g_n,
\qquad n\geq1.
\]

We now show that every $h\in\mathcal H_K$ can be approximated, in the RKHS norm, by elements of $\mathcal H_K\cap\ell_1$. Fix an arbitrary $h\in\mathcal H_K$ and consider its image
\[
Y=\mathcal{I}h\in\mathcal G.
\]
Since $\mathcal G$ is the closed linear span of the random variables $g_n$, the random variable $Y$ can be approximated in $L^2(\Omega)$ by finite linear combinations of the components of the process $g$. The construction below exploits, in addition, the almost-sure stability of the entire sample path $g$.

For each $m\geq1$, define the event
\[
A_m
=
\left\{
|Y|\leq m,\;
\|g\|_1\leq m
\right\}.
\]
Thus, $A_m$ contains the outcomes for which both $Y$ and the $\ell_1$-norm of the sample path are bounded by $m$. Define the truncated random variable
\[
Z_m
=
Y\mathbf 1_{A_m},
\]
where $\mathbf 1_{A_m}$ denotes the indicator function of $A_m$.

Since $Y\in L^2(\Omega)$, it is finite almost surely, while by assumption $\|g\|_1<\infty$ almost surely. Hence, as $m$ increases, the events $A_m$ increase to an event of probability one:
\[
A_m\uparrow
\left\{
|Y|<\infty,\;
\|g\|_1<\infty
\right\}.
\]
Consequently,
\[
\mathbf 1_{A_m}\longrightarrow 1
\qquad\text{almost surely},
\]
and therefore
\[
Z_m\longrightarrow Y
\qquad\text{almost surely}.
\]
Moreover,
\[
|Z_m-Y|^2
=
|Y|^2|\mathbf 1_{A_m}-1|^2
\leq |Y|^2,
\]
and $|Y|^2\in L^1(\Omega)$. The dominated convergence theorem then gives
\[
\mathbb E|Z_m-Y|^2\longrightarrow0,
\]
that is,
\[
Z_m\longrightarrow Y
\qquad\text{in }L^2(\Omega).
\]

The truncated variable $Z_m$ need not belong to $\mathcal G$, so we project it onto this space and define
\[
Y_m=P_{\mathcal G}Z_m,
\]
where $P_{\mathcal G}$ is the orthogonal projection of $L^2(\Omega)$ onto $\mathcal G$. Since $Y\in\mathcal G$, we have $P_{\mathcal G}Y=Y$. Moreover, %an orthogonal projection is a contraction, and hence
\[
\begin{aligned}
\|Y_m-Y\|_{L^2}
&=
\|P_{\mathcal G}Z_m-P_{\mathcal G}Y\|_{L^2}\\
&=
\|P_{\mathcal G}(Z_m-Y)\|_{L^2}\\
&\leq
\|Z_m-Y\|_{L^2}.
\end{aligned}
\]
It follows that
\[
Y_m\longrightarrow Y
\qquad\text{in }L^2(\Omega).
\]

We now map $Y_m$ back to the RKHS by setting
\[
h_m=\mathcal{I}^{-1}Y_m\in\mathcal H_K.
\]
Now, we show that  $h_m\in\ell_1$. For every $n\geq1$, the reproducing property and the isometry $\mathcal{I}$ give

\begin{eqnarray}\nonumber
h_m(n)
&=&
\langle h_m,K(\cdot,n)\rangle_{\mathcal H_K}\\ \label{chainInner}
&=&
\langle Y_m,g_n\rangle_{L^2(\Omega)}.
\end{eqnarray}
To evaluate the last inner product, recall that
\[
Y_m=P_{\mathcal G}Z_m,
\]
so that
\[
Z_m-Y_m\perp\mathcal G.
\]
Since $g_n\in\mathcal G$, it follows that
\[
\langle Z_m-Y_m,g_n\rangle_{L^2(\Omega)}=0,
\]
and therefore from \eqref{chainInner} we obtain
\[
\langle Y_m,g_n\rangle_{L^2(\Omega)}
=
\langle Z_m,g_n\rangle_{L^2(\Omega)}.
\]
Thus,
\[
h_m(n)
=
\mathbb E[Z_m g_n].
\]

The use of the triangle inequality gives
\[
|h_m(n)|
\leq
\mathbb E\!\left[|Z_m|\,|g_n|\right].
\]
Summing over $n$ and applying Tonelli's theorem, which allows interchange of expectation and an infinite sum of nonnegative random variables, yields
\begin{align*}
\sum_{n=1}^{\infty}|h_m(n)|
&\leq
\sum_{n=1}^{\infty}
\mathbb E\!\left[|Z_m|\,|g_n|\right]\\
&=
\mathbb E\!\left[
|Z_m|\sum_{n=1}^{\infty}|g_n|
\right].
\end{align*}
Now $Z_m$ vanishes outside $A_m$, while on $A_m$ we have both
\[
|Z_m|=|Y|\leq m
\qquad\text{and}\qquad
\|g\|_1\leq m.
\]
Therefore
\[
\mathbb E\!\left[
|Z_m|\sum_{n=1}^{\infty}|g_n|
\right]
\leq m^2.
\]
and hence
\[
\sum_{n=1}^{\infty}|h_m(n)|<\infty.
\]
Thus
\[
h_m\in\mathcal H_K\cap\ell_1.
\]

Finally, since $\mathcal{I}$ is an isometry,
\[
\|h_m-h\|_{\mathcal H_K}
=
\|Y_m-Y\|_{L^2}
\longrightarrow0.
\]
Since $h\in\mathcal H_K$ was arbitrary, we have constructed, for every $h$, a sequence $\{h_m\}\subset\mathcal H_K\cap\ell_1$ converging to $h$ in the RKHS norm. This proves
\eqref{eq:density}.

We now prove sufficiency. We discuss it for the infinite-dimensional case; the finite-dimensional case follows identically by replacing the infinite basis and sums below with their finite counterparts.
Assume that \eqref{eq:density} holds. Since $\mathcal H_K$ is separable and $\mathcal H_K\cap\ell_1$ is dense in $\mathcal H_K$, we can choose a countable dense subset of $\mathcal H_K\cap\ell_1$. Applying the Gram--Schmidt procedure to this set, after discarding any linearly dependent elements, yields an orthonormal basis
\[
\{\phi_r\}_{r\geq1}
\]
of $\mathcal H_K$ such that
\[
\phi_r\in\ell_1,
\qquad r\geq1.
\]
Indeed, every vector produced by Gram--Schmidt is a finite linear combination of elements of $\mathcal H_K\cap\ell_1$ and therefore still belongs to $\ell_1$.

We now construct explicitly a stochastic process having covariance $K$ and stable sample paths. Let $(p_r)_{r\geq1}$ be any sequence satisfying
\[
p_r>0,
\qquad
\sum_{r=1}^{\infty}p_r=1,
\]
and introduce a random variable $J$ taking values on $\mathbb N$ with probability mass function
\[
\mathbb P(J=r)=p_r,
\qquad r\geq1.
\]
Let $\varepsilon$ be a Rademacher random variable, independent of $J$, namely
\[
\mathbb P(\varepsilon=1)
=
\mathbb P(\varepsilon=-1)
=
\frac12.
\]
Define the random sequence
\begin{equation}
    g
    =
    \frac{\varepsilon}{\sqrt{p_J}}\phi_J.
    \label{eq:singleatomprior}
\end{equation}

For every realization, $J$ takes some value $r\in\mathbb N$, so that
\[
g
=
\frac{\varepsilon}{\sqrt{p_r}}\phi_r.
\]
Since $\phi_r\in\ell_1$ and $p_r>0$,
\[
\|g\|_1
=
\frac{\|\phi_r\|_1}{\sqrt{p_r}}
<
\infty.
\]
Hence
\[
g\in\ell_1
\qquad\text{almost surely}.
\]
%in fact for every realization.

Moreover, since $\varepsilon$ is independent of $J$ and
$\mathbb E[\varepsilon]=0$, we have
\[
\mathbb E[g_n]=0,
\qquad n\geq1.
\]
It remains to verify the covariance. For $i,j\geq1$,
\begin{align*}
\mathbb E[g_i g_j]
&=
\sum_{r=1}^{\infty}
p_r
\frac{1}{p_r}
\phi_r(i)\phi_r(j)\\
&=
\sum_{r=1}^{\infty}
\phi_r(i)\phi_r(j).
\end{align*}
To identify the last series, expand the kernel section $K(\cdot,j)$ in the orthonormal basis $\{\phi_r\}$:
\[
K(\cdot,j)
=
\sum_{r=1}^{\infty}
\langle K(\cdot,j),\phi_r\rangle_{\mathcal H_K}\phi_r
=
\sum_{r=1}^{\infty}
\phi_r(j)\phi_r,
\]
where the second equality follows from the reproducing property. Evaluating this identity at $i$ gives
\[
K(i,j)
=
\sum_{r=1}^{\infty}
\phi_r(i)\phi_r(j).
\]
Therefore
\[
\mathbb E[g_i g_j]=K(i,j),
\qquad i,j\geq1.
\]
The process $g$ is thus zero mean, has covariance $K$, and has sample paths in $\ell_1$ almost surely. Hence $K$ is stability-compatible.
\end{proof}

The theorem shows that full RKHS stability is not necessary. It is enough that stable elements form a dense ``skeleton'' of $\mathcal H_K$: unstable functions may belong to the RKHS, while every RKHS element can still be approximated arbitrarily well in the RKHS norm by stable ones.

\section{Sufficient conditions for stability compatibility}
\label{sec:sufficient}

\subsection{General kernel conditions}

As an immediate consequence, the classical RKHS-stability condition is sufficient for stability-compatibility.

\begin{corollary}
\label{cor:stableimplies}
Every RKHS-stable kernel is stability-compatible.
\end{corollary}

%This corollary is also implied by the more general sample-path construction of \citep{LukicBeder2001} but Theorem~\ref{thm:main} goes far beyond by identifying the exact condition when the target path space is $\ellone$.

Theorem~\ref{thm:main} also yields simple kernel-level sufficient conditions.

\begin{proposition}
\label{prop:sections}
Let $K$ be a positive semidefinite kernel on $\mathbb N$. Each of the following conditions is sufficient for stability-compatibility:
\begin{enumerate}
    \item every kernel section is absolutely summable, namely
    \begin{equation}
        K(\cdot,n)\in\ellone
        \qquad\text{for every }n\in\mathbb N;
        \label{eq:sections-l1}
    \end{equation}

    \item there exists a sequence $\{w_h\}_{h\ge1}$ with $w_h>0$ such that the principal Gram matrices
    \begin{equation}
        K_N=[K(i,j)]_{i,j=1}^N
        \label{eq:KN-def}
    \end{equation}
    satisfy
    \begin{equation}
        K_N\succeq W_N,
        \qquad
        W_N:=\operatorname{diag}(w_1,\ldots,w_N),
        \qquad N\ge1,
        \label{eq:weighted-lower-gram}
    \end{equation}
    and, for every kernel section,
    \begin{equation}
        \sum_{h=1}^{\infty}
        \frac{|K(h,n)|^2}{w_h}<\infty,
        \qquad n\in\mathbb N.
        \label{eq:weighted-sections}
    \end{equation}
\end{enumerate}
\end{proposition}

\begin{proof}
Under condition~1, every finite linear combination of kernel sections belongs to $\HK\cap\ellone$. Their span is dense in $\HK$ by \eqref{eq:sectionspan}, so Theorem~\ref{thm:main} gives stability-compatibility. Condition~2 is proved in \ref{app:lower-gram}.
\end{proof}

Condition~2 contains the following simpler uniform criterion as the special case $w_h = m>0$: then \eqref{eq:weighted-lower-gram} becomes $K_N\succeq mI_N$, while \eqref{eq:weighted-sections} is equivalent to requiring every kernel section to belong to $\elltwo$. 

\medskip

We next give some simple examples illustrating the two conditions and showing that they capture different mechanisms leading to stability-compatibility.

\paragraph{Diagonal kernels}
Every diagonal kernel
\[
K(i,j)=\lambda_i\delta_{ij},
\qquad \lambda_i\ge0,
\]
is stability-compatible, independently of the behavior of the sequence $\{\lambda_i\}$. Indeed, every kernel section has finite support and therefore belongs to $\ellone$, so condition~1 applies. In particular, stability-compatibility may hold even when the diagonal does not decay, or even grows with the index.

\paragraph{Gaussian/RBF kernel}
For
\[
K(i,j)=\exp\{-\alpha(i-j)^2\},\qquad \alpha>0,
\]
every kernel section belongs to $\ellone$, so condition~1 gives stability-compatibility. On the other hand, $K(i,i)=1$ for every $i$, and the finite-trace necessary condition for RKHS stability fails \citep{BisiaccoPillonetto2020}. Thus the RBF kernel is neither RKHS-stable nor Gaussian-stable.

\paragraph{Stationary ARMA covariances}
Consider a causal stationary ARMA$(p,q)$ process
\begin{equation}
x_t-a_1x_{t-1}-\cdots-a_px_{t-p}
=
e_t+b_1e_{t-1}+\cdots+b_qe_{t-q},
\label{eq:example-arma}
\end{equation}
where the autoregressive polynomial is stable. Its autocovariance sequence is absolutely summable, so condition~1 gives stability-compatibility, independently of possible zeros of the moving-average polynomial on the unit circle. Since $K(i,i)=r(0)>0$ for every $i$, the finite-trace condition fails, and the kernel is neither RKHS-stable nor Gaussian-stable.

\paragraph{A kernel satisfying condition~2 but not condition~1}
Let $u_n=1/n$ and
\begin{equation}
K(i,j)=\delta_{ij}+u_i u_j.
\label{eq:rankone-example}
\end{equation}
Since $u\in\elltwo\setminus\ellone$, each section
\[
K(\cdot,j)=e_j+u_j u
\]
belongs to $\elltwo$ but not to $\ellone$. Moreover,
\[
K_N=I_N+u_Nu_N^\top\succeq I_N
\]
for every $N$. Thus condition~2 applies with $w_h = 1$, whereas condition~1 does not. Since the kernel sections themselves are not in $\ellone$, the kernel is neither RKHS-stable nor Gaussian-stable.

\subsection{A spectral sufficient condition for stationary kernels}

Consider a stationary kernel
\[
K(i,j)=r(i-j),
\qquad i,j\in\mathbb N,
\]
where $r:\mathbb Z\to\mathbb R$ is its covariance sequence. By
Herglotz's theorem,
\begin{equation}\label{H1}
r(k)
=
\int_{-\pi}^{\pi}
e^{\mathrm{i}k\omega}\,d\mu(\omega),
\qquad k\in\mathbb Z,
\end{equation}
for a finite positive spectral measure $\mu$  \citep{LindquistPicci2015}.
When the spectral measure is absolutely continuous, we use 
the normalization
\begin{equation}\label{H2}
d\mu(\omega)
=
f(\omega)\frac{d\omega}{2\pi}.
\end{equation}
With this convention, the classical Szeg\H{o} condition is
\begin{equation}\label{eq:szego-main}
\int_{-\pi}^{\pi}
\log f(\omega)\frac{d\omega}{2\pi}
>
-\infty,
\end{equation}
which, in prediction-theoretic terms, is equivalent to a strictly positive one-step innovation variance \citep{Bingham2012}.

\begin{proposition}[Stationary spectral criterion]
\label{prop:stationary-szego-compatible}
Let \(K(i,j)=r(i-j)\) be a stationary positive semidefinite kernel. If its spectral measure admits a spectral density \(f\) satisfying the Szeg\H{o} condition \eqref{eq:szego-main}, then \(K\) is stability-compatible.
\end{proposition}

The proof, given in \ref{app:stationary-szego}, uses the same canonical isometry
\[
\mathcal I K(\cdot,n)=g_n
\]
as in the proof of Theorem~\ref{thm:main}. Under this identification, the Szeg\H{o} condition becomes nondegeneracy of the RKHS counterpart of the one-step innovation, while absolute continuity is equivalent, under Szeg\H{o}, to
\[
\bigcap_{N\ge1}
\overline{\operatorname{span}\bigl\{K(\cdot,n):n\ge N\bigr\}}^{\,\mathcal H_K}
=
\{0\}.
\]

For stationary kernels, a simple way to verify condition~2 of Proposition~\ref{prop:sections} is through the uniform spectral bounds
\[
0<m\le f(\omega)\le M<\infty
\qquad\text{a.e.}
\]
Indeed, by the classical Grenander--Szeg\H{o} theory for Toeplitz matrices, the eigenvalues of the Gram matrices $K_N$ are confined to the essential range of the generating spectral density \citep{GrenanderSzego1958}. Hence the lower bound on $f$ gives
\[
K_N\succeq mI_N
\qquad\text{for every }N,
\]
so that one may take $w_h\equiv m$ in \eqref{eq:weighted-lower-gram}. The upper bound implies that every kernel section belongs to $\ell_2$.\footnote{Since $f\le M$ on $[-\pi,\pi]$, one has $f\in L^2[-\pi,\pi]$. By Parseval's identity, its Fourier coefficients $r(k)$ therefore belong to $\ell_2(\mathbb Z)$. Since $K(i,j)=r(i-j)$, it follows that $K(\cdot,j)\in\ell_2$ for every $j$.}
Therefore \eqref{eq:weighted-sections} also holds. The variable weights in condition~2 make this criterion more flexible than the uniform bound above. The next example shows, however, that even the weighted diagonal criterion may fail while the Szeg\H{o} criterion still applies.

\paragraph{A spectral example beyond Proposition~\ref{prop:sections}}
Consider the spectral density
\begin{equation}
    f_\rho(\omega)
    =
    |\omega|+\rho\,\mathbf 1_{\{|\omega|>\pi/2\}},
    \qquad \rho>0.
    \label{eq:spectral-only-example}
\end{equation}
It is positive almost everywhere and satisfies the Szeg\H{o} condition,
since its only zero is at $\omega=0$ and $\log|\omega|$ is locally
integrable. Hence Proposition~\ref{prop:stationary-szego-compatible}
gives stability-compatibility.

For $k\ne0$, using \eqref{H1} and \eqref{H2}, the covariance
coefficients are
\begin{equation}
    r(k)
    =
    \frac{(-1)^k-1}{\pi k^2}
    -
    \frac{\rho}{\pi k}
    \sin\!\left(\frac{\pi k}{2}\right).
    \label{eq:spectral-only-covariance}
\end{equation}
The second term is of order $1/|k|$ along the odd integers, so
$r\notin\ellone$. Thus Condition~1 in
Proposition~\ref{prop:sections} fails.

We now show that Condition~2 fails for every choice of positive weights.
Assume that \eqref{eq:weighted-lower-gram} holds and let
$\mathbf 1_N=(1,\ldots,1)^\top$. Then
\[
\sum_{h=1}^N w_h
\le
\mathbf 1_N^\top K_N\mathbf 1_N.
\]
Using the spectral representation,
\begin{align*}
\mathbf 1_N^\top K_N\mathbf 1_N
&=
\sum_{p,q=1}^N r(p-q)\\
&=
\int_{-\pi}^{\pi}
f_\rho(\omega)
\left|
\sum_{p=1}^N e^{\mathrm{i}p\omega}
\right|^2
\frac{d\omega}{2\pi}\\
&=
N\int_{-\pi}^{\pi}
f_\rho(\omega)F_N(\omega)\frac{d\omega}{2\pi},
\end{align*}
where
\[
F_N(\omega)
=
\frac1N
\left|
\sum_{p=1}^N e^{\mathrm{i}p\omega}
\right|^2
\]
is the Fej\'er kernel \citep{GrenanderSzego1958}. The standard bound
\[
F_N(\omega)
\le
C_F\min\!\left\{
N,\frac{1}{N\omega^2}
\right\}
\]
and
\[
f_\rho(\omega)
\le
C_\rho|\omega|,
\qquad
C_\rho=1+\frac{2\rho}{\pi},
\]
lead to
\begin{align*}
\sum_{h=1}^N w_h
&\le
N\int_{-\pi}^{\pi}
f_\rho(\omega)F_N(\omega)\frac{d\omega}{2\pi}\\
&\le
\frac{NC_\rho}{2\pi}
\int_{-\pi}^{\pi}
|\omega|F_N(\omega)\,d\omega\\
&=
O(\log N).
\end{align*}
Hence, for some constant $C>0$,
\begin{equation}
\sum_{h=1}^N w_h
\le
C\log N
\label{eq:spectral-only-weight-growth}
\end{equation}
for all sufficiently large $N$. On the other hand, for odd $k$, \eqref{eq:spectral-only-covariance}
implies
\[
|r(k)|
\ge
\frac{\rho}{\pi|k|}
-
\frac{2}{\pi k^2}.
\]
Thus there exist $c_0>0$ and an even integer $h_0$ such that
\[
|K(h,1)|
=
|r(h-1)|
\ge
\frac{c_0}{h},
\qquad
h\ge h_0,\quad h\ {\rm even}.
\]
By Cauchy--Schwarz and \eqref{eq:spectral-only-weight-growth},
\[
\left(
\sum_{\substack{h_0\le h\le N\\ h\ {\rm even}}}
\frac1h
\right)^2
\le
C\log N
\sum_{\substack{h_0\le h\le N\\ h\ {\rm even}}}
\frac{1}{h^2w_h}.
\]
Since
\[
\sum_{\substack{h_0\le h\le N\\ h\ {\rm even}}}
\frac1h
\asymp
\log N,
\]
there exists $c_1>0$ such that
\[
\sum_{\substack{h_0\le h\le N\\ h\ {\rm even}}}
\frac{1}{h^2w_h}
\ge
c_1\log N
\]
for all sufficiently large $N$. Hence
\[
\sum_{\substack{h\ge h_0\\ h\ {\rm even}}}
\frac{1}{h^2w_h}
=
\infty.
\]
Finally,
\[
\sum_{h=1}^{\infty}
\frac{|K(h,1)|^2}{w_h}
\ge
c_0^2
\sum_{\substack{h\ge h_0\\ h\ {\rm even}}}
\frac{1}{h^2w_h}
=
\infty,
\]
contradicting \eqref{eq:weighted-sections}. Thus Condition~2 of
Proposition~\ref{prop:sections} fails for every sequence of positive
weights.

\section{A strict hierarchy of kernel stability notions}
\label{sec:hierarchy}

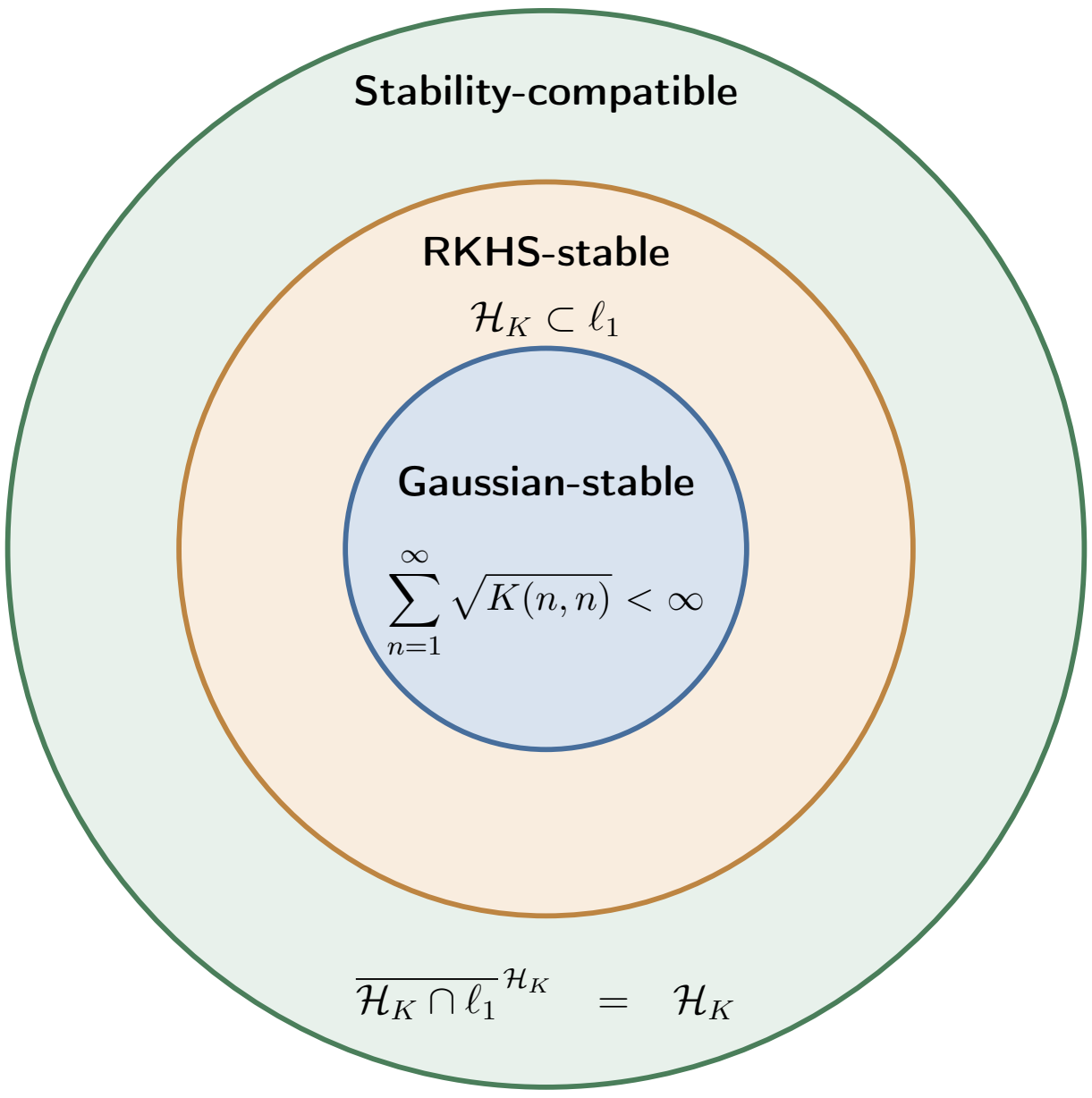
\begin{figure}[t]
\centering

\resizebox{0.5\columnwidth}{!}{%
\begin{tikzpicture}[font=\sffamily]

\definecolor{compatgreen}{RGB}{92,157,112}
\definecolor{rkhsorange}{RGB}{222,157,78}
\definecolor{gaussblue}{RGB}{84,130,184}

\filldraw[
    fill=compatgreen!14,
    draw=compatgreen!80!black,
    line width=1.5pt
] (0,0) circle (5.5cm);

\filldraw[
    fill=rkhsorange!18,
    draw=rkhsorange!85!black,
    line width=1.5pt
] (0,0) circle (3.75cm);

\filldraw[
    fill=gaussblue!22,
    draw=gaussblue!85!black,
    line width=1.5pt
] (0,0) circle (2.05cm);

\node[
    font=\sffamily\bfseries\large,
    align=center
] at (0,4.65)
{Stability-compatible};

\node[
    font=\sffamily\large,
    align=center,
    text width=8.5cm
] at (0,-4.55)
{$\displaystyle
\overline{\mathcal H_K\cap\ell_1}^{\,\mathcal H_K}
=
\mathcal H_K$};

\node[
    font=\sffamily\bfseries\large,
    align=center
] at (0,3.05)
{RKHS-stable};

\node[
    font=\sffamily\large,
    align=center
] at (0,2.35)
{$\mathcal H_K\subset\ell_1$};

\node[
    font=\sffamily\bfseries\large,
    align=center
] at (0,0.70)
{Gaussian-stable};

\node[
    font=\sffamily\normalsize,
    align=center,
    text width=3.6cm
] at (0,-0.55)
{$\displaystyle
\sum_{n=1}^{\infty}\sqrt{K(n,n)}
<\infty$};

\end{tikzpicture}%
}

\caption{\textbf{Strict hierarchy of kernel stability notions.}
The three nested disks represent classes of positive semidefinite kernels.
The class of Gaussian-stable kernels is strictly contained in the class of
RKHS-stable kernels, which is in turn strictly contained in the class of
stability-compatible kernels.}
\label{fig:hierarchy}
\end{figure}

The notions introduced in Section~\ref{sec:preliminaries} satisfy
\[
\begin{gathered}
\text{Gaussian stability}\\
\Downarrow\\
\text{RKHS stability}\\
\Downarrow\\
\text{stability-compatibility}.
\end{gathered}
\]
The first implication follows from the Gaussian stability condition discussed
above, whereas the second follows immediately from
Theorem~\ref{thm:main}. Hence the three notions define nested classes of
kernels, as illustrated in Fig.~\ref{fig:hierarchy}. The strictness of the
second inclusion has already been established by the RBF example in
Section~\ref{sec:sufficient}. To show that the first inclusion is also strict,
consider the diagonal kernel 
$$
K(i,j)=\lambda_i\delta_{ij}, \quad \lambda_i=i^{-2}.
$$ 
It is absolutely summable and hence RKHS-stable
\citep{BisiaccoPillonetto2020}. On the other hand,
$$\sum_{i=1}^{\infty}\sqrt{K(i,i)}
=\sum_{i=1}^{\infty}\frac{1}{i}=\infty.
$$
Therefore the DSRI condition fails, and the Gaussian process with covariance
$K$ does not have sample paths in $\ell_1$ almost surely.

%%%%%%%%%%%%%%%%%% FIGURE %%%%%%%%%%%%%%%%%%%%%%%%%%

%%%%%%%%%%%%%%%%%% END OF FIGURE %%%%%%%%%%%%%%%%%%%%%%%%%%

%%%%%%%%%%%%%%%%%%%%%
\section{A constructive family of stable non-Gaussian priors}
\label{sec:twoscale}

The single-atom construction used in the proof of Theorem~\ref{thm:main}
shows that condition~\eqref{eq:density} is sufficient for the existence of a
stable stochastic process with covariance $K$.  Its role is mainly
constructive, however: each realization activates only one RKHS direction.
We now show that the same condition supports a substantially richer class of
priors in which infinitely many RKHS coefficients may be nonzero, while only
finitely many large-scale activations occur almost surely. The covariance is
still exactly $K$ and the impulse response remains in $\ellone$ almost
surely.

Assume throughout this section that $K$ is stability-compatible.  By
Theorem~\ref{thm:main}, $\HK\cap\ellone$ is dense in $\HK$.  Since $\HK$ is
separable, Gram--Schmidt applied to a countable dense subset of
$\HK\cap\ellone$ yields an orthonormal basis
\begin{equation}
    \{\phi_j\}_{j\geq1}\subset \HK\cap\ellone .
    \label{eq:stable-onb}
\end{equation}
Thus the following construction applies to every kernel satisfying the
necessary and sufficient condition~\eqref{eq:density}.

\begin{proposition}[Two-scale stable prior]
\label{prop:twoscale}
Let $\{\phi_j\}_{j\geq1}$ be an orthonormal basis satisfying
\eqref{eq:stable-onb}.  Let $\{Z_j\}_{j\geq1}$ be independent standard Gaussian
random variables and let $\{B_j\}_{j\geq1}$ be independent Bernoulli random
variables, independent of the $Z_j$, with
\begin{equation}
    \mathbb P(B_j=1)=q_j,\qquad 0<q_j<1,
    \qquad
    \sum_{j=1}^{\infty} q_j<\infty.
    \label{eq:twoscale-q}
\end{equation}
Choose numbers $c_j$ such that
\begin{equation}
    0<c_j\leq1,
    \qquad
    \sum_{j=1}^{\infty} c_j\|\phi_j\|_1<\infty,
    \label{eq:twoscale-c}
\end{equation}
and define
\begin{equation}
    d_j^2
    =
    \frac{1-(1-q_j)c_j^2}{q_j}.
    \label{eq:twoscale-d}
\end{equation}
Set
\begin{equation}
    S_j=(1-B_j)c_j+B_jd_j,
    \qquad
    a_j=S_jZ_j,
    \label{eq:twoscale-coeff}
\end{equation}
and
\begin{equation}
    g=\sum_{j=1}^{\infty} a_j\phi_j.
    \label{eq:twoscale-prior}
\end{equation}
Then the series in \eqref{eq:twoscale-prior} converges absolutely in
$\ellone$ almost surely.  The resulting process is zero mean, has covariance
$K$, and therefore satisfies
\begin{equation}
    \mathbb P(g\in\ellone)=1.
\end{equation}
For each $j$, the marginal law of $a_j$ is the centered two-Gaussian mixture
\begin{equation}
    a_j\sim
    (1-q_j)\,\mathcal N(0,c_j^2)
    +q_j\,\mathcal N(0,d_j^2).
    \label{eq:twoscale-mixture}
\end{equation}
\end{proposition}

\begin{proof}
Because \(\sum_j q_j<\infty\), the first Borel--Cantelli lemma implies that
only finitely many events \(\{B_j=1\}\) occur almost surely.  Hence the
contribution of the large-scale terms,
\[
    \sum_{j:B_j=1} d_j Z_j\phi_j,
\]
is almost surely a finite linear combination of elements of $\ellone$.

For the remaining terms, Tonelli's theorem and \eqref{eq:twoscale-c} give
\begin{align*}
\mathbb E\!\left[
    \sum_{j=1}^{\infty} c_j|Z_j|\,\|\phi_j\|_1
\right]
&=
\mathbb E|Z_1|
\sum_{j=1}^{\infty} c_j\|\phi_j\|_1\\
&<\infty.
\end{align*}
Therefore
\[
    \sum_{j=1}^{\infty} c_j|Z_j|\,\|\phi_j\|_1<\infty
    \qquad\text{almost surely}.
\]
Combining the two parts shows that \eqref{eq:twoscale-prior} converges
absolutely in $\ellone$ almost surely.

Each coefficient is centered and, by construction,
\begin{align*}
    \mathbb E[a_j^2]
    &=
    \mathbb E[S_j^2]\mathbb E[Z_j^2]\\
    &=(1-q_j)c_j^2+q_jd_j^2 =1.
\end{align*}
Moreover, $\mathbb E[a_r a_s]=\delta_{rs}$. Since
\[
\sum_{r=1}^{\infty}\phi_r(n)^2=K(n,n)<\infty,
\]
the series defining each component converges in $L^2$. Since the
$\ellone$ convergence established above also implies coordinatewise
almost-sure convergence, its $L^2$ limit is precisely $g_n$. In
particular, $\mathbb E[g_n]=0$ for every $n$, and
\[
\mathbb E[g_i g_j]
=
\sum_{r=1}^{\infty}\phi_r(i)\phi_r(j)
=
K(i,j),
\]
where the last equality is the orthonormal-basis expansion of the kernel.
Thus $g$ is zero mean, has covariance $K$, and has stable sample paths
almost surely. Finally, \eqref{eq:twoscale-mixture} follows immediately from
\eqref{eq:twoscale-coeff}.
%Finally, since $B_j=0$ with probability $1-q_j$ and
%$B_j=1$ with probability $q_j$, we have
%\[
%a_j\sim
%(1-q_j)\mathcal N(0,c_j^2)
%+
%q_j\mathcal N(0,d_j^2),
%\]
%which proves \eqref{eq:twoscale-mixture}.
%Thus $g$ is zero mean, has covariance $K$, and has stable sample paths
%almost surely.
%Moreover, independence and zero means imply
%\[
%    \mathbb E[a_i a_j]=\delta_{ij}.
%\]
%For every fixed $n$,
%\[
%\sum_{j=1}^{\infty}
%\mathbb E[a_j^2]\phi^2_j(n)
%=
%\sum_{j=1}^{\infty}\phi^2_j(n)
%=
%K(n,n)
%<\infty.
%\]
%Hence the series
%\[
%\sum_{j=1}^{\infty}a_j\phi_j(n)
%\]
%converges in $L^2$. Since the series in
%\eqref{eq:twoscale-prior} also converges almost surely in $\ellone$,
%this $L^2$ limit is precisely $g_n$. Therefore, using
%$\mathbb E[a_r a_s]=\delta_{rs}$,
%\[
%\mathbb E[g_i g_j]
%=
%\sum_{r=1}^{\infty}\phi_r(i)\phi_r(j)
%=
%K(i,j),
%\]
%where the last equality follows from the orthonormal-basis expansion
%of the kernel. Thus $g$ has covariance $K$ and stable sample paths
%almost surely.
\end{proof}

The construction separates second-order geometry from higher-order prior
structure.  Equation~\eqref{eq:twoscale-d} forces every RKHS expansion coefficient $a_j$ in \eqref{eq:twoscale-mixture} to
have unit variance, so changing $q_j$ and $c_j$ does not alter the covariance
kernel.  At the same time, these quantities strongly affect the marginal
distributions: when $q_j$ is small, most draws come from the narrow Gaussian
component with scale $c_j$, whereas rare draws from the wider component with
scale $d_j$ restore the prescribed second moment.  The resulting prior is
therefore non-Gaussian unless $c_j=d_j$ for every $j$.

A convenient parameterization, used throughout the numerical study, is obtained by fixing
\begin{equation}
    q_j=\frac{1}{(j+1)^2},
    \qquad
    c_j=\frac{\exp(-\beta j)}{1+\|\phi_j\|_1},
    \qquad \beta>0.
    \label{eq:twoscale-param}
\end{equation}
Thus no additional parameter controlling the activation probabilities is introduced.
The sequence $\{q_j\}$ is summable and, by independence,
\[
    \mathbb P(B_j=0\ \text{for all }j)
    =
    \prod_{j=1}^{\infty}
    \left(1-\frac{1}{(j+1)^2}\right)
    =
    \frac12.
\]
Equivalently,
\[
    \mathbb P(B_j=1\ \text{for at least one }j)=\frac12,
\]
so the rare, large-scale component remains nondegenerate without introducing
an additional tuning parameter. Moreover,
\[
    \sum_{j=1}^{\infty}c_j\|\phi_j\|_1
    \leq
    \sum_{j=1}^{\infty}e^{-\beta j}
    <\infty,
\]
so Proposition~\ref{prop:twoscale} applies to any stability-compatible kernel.

If an overall covariance scale $\lambda>0$ is introduced by replacing $g$
with $\sqrt{\lambda}\,g$, the resulting prior has covariance $\lambda K$.
Thus, for a fixed kernel $K$ and with the sequence $\{q_j\}$ chosen as above,
the prior depends only on the overall covariance scale $\lambda$ and on
$\beta$, which controls the decay of the ordinary, small-scale component.
%has the stable RBF model introduced below contains three hyperparameters: the RBF shape parameter $\alpha$, the overall covariance scale $\lambda$, and the parameter $\beta$ controlling the decay of the ordinary, small-scale component.

\subsection{A stable non-Gaussian RBF prior}
\label{sec:stable-rbf}

The construction above becomes especially interesting for the RBF kernel
\begin{equation}
    K_{\rm RBF}(i,j)
    =
    \exp\!\left(-\alpha(i-j)^2\right),
    \qquad \alpha>0.
    \label{eq:rbf-kernel-stable}
\end{equation}
Since
\[
K_{\rm RBF}(n,n)=1,
\]
the corresponding Gaussian prior cannot have stable sample paths.
However, the RBF kernel is stability-compatible, so
Proposition~\ref{prop:twoscale} yields a non-Gaussian prior with the same
covariance and $\mathbb P(g\in\ellone)=1$.
We refer to this model as the \emph{stable RBF prior}.

Its interpretation differs from that of classical stable kernels such as TC.
For
\[
K_{\rm TC}(i,j)
=
\lambda_{\rm TC}\beta_{\rm TC}^{\max(i,j)},
\qquad
0<\beta_{\rm TC}<1,
\]
both temporal dependence and decay are encoded directly in the covariance,
whose marginal variance decreases exponentially with the lag. The TC model
therefore has two hyperparameters, $\lambda_{\rm TC}$ and $\beta_{\rm TC}$.

By contrast, the stable RBF prior has hyperparameters
$\alpha$, $\lambda$, and $\beta$: the covariance controls second-order
dependence and smoothness, while almost-sure stability is enforced by the
higher-order non-Gaussian law. Thus stable sample paths are obtained even
though the pointwise variance remains constant.

\begin{figure*}[t]
\centering
\includegraphics[scale=0.34]{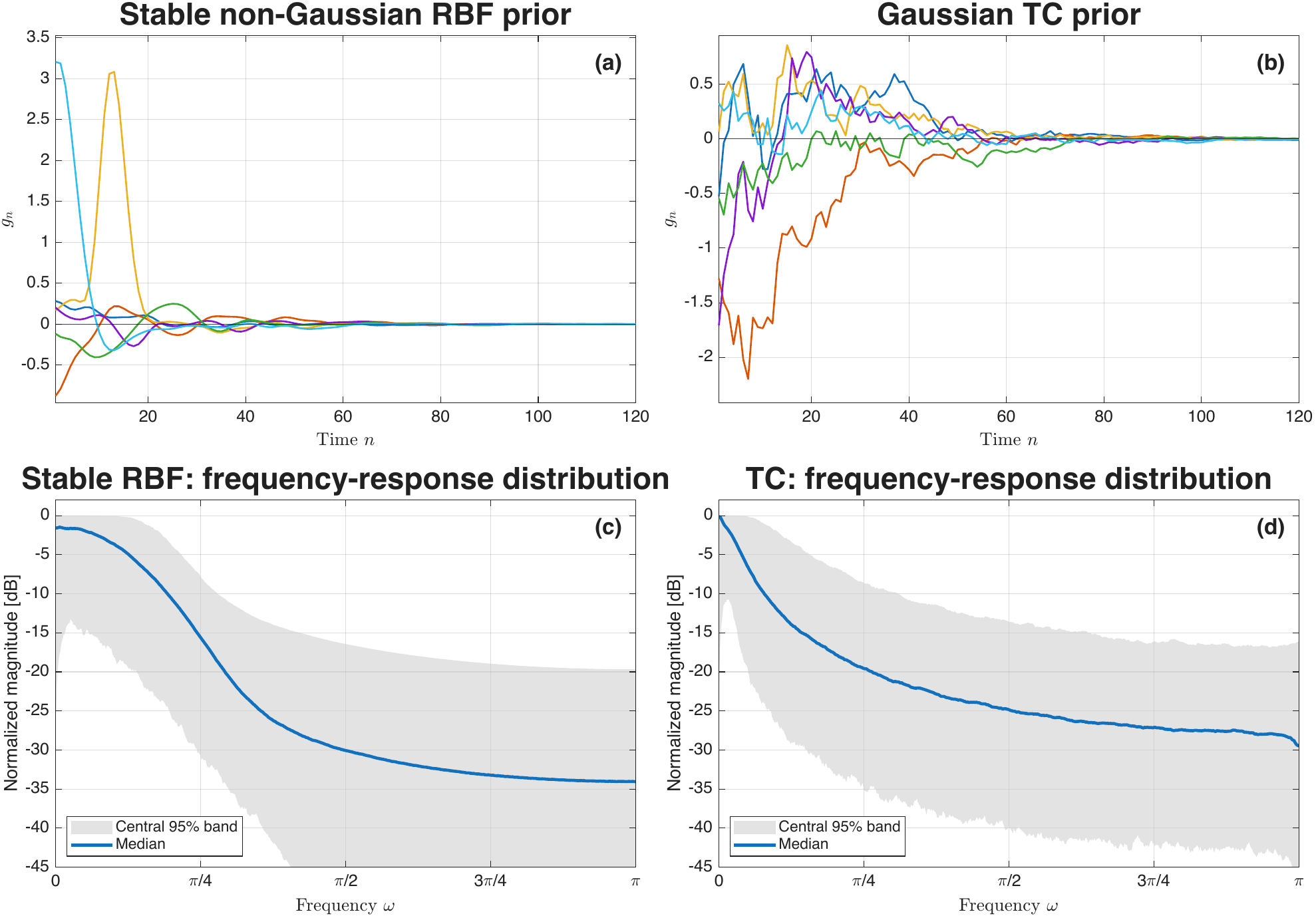}
\caption{\textbf{Stable non-Gaussian RBF prior versus Gaussian TC prior.}
\textbf{a}, Sample impulse responses generated by the stable RBF prior.
\textbf{b}, Sample impulse responses generated by the Gaussian TC prior.
The shape parameters are selected so that the two models have the same correlation between adjacent coefficients.
\textbf{c}, Distribution of the normalized frequency-response magnitude for the stable RBF prior.
\textbf{d}, Corresponding distribution for the TC prior.
Solid lines denote the median and shaded regions the central $95\%$ probability band.}
%The two priors enforce stability through fundamentally different mechanisms. In the TC model, decay is already encoded in the covariance through its decreasing diagonal. In the stable RBF model, the covariance retains the stationary RBF form with constant diagonal, while absolute summability is instead enforced by the non-Gaussian two-scale distribution of the expansion coefficients.}
\label{fig:stable-rbf-vs-tc}
\end{figure*}

Figure~\ref{fig:stable-rbf-vs-tc} illustrates this distinction. To compare the two priors at a similar level of local dependence, we impose the constraint
\begin{equation}
    e^{-\alpha}
    =
    \sqrt{\beta_{\rm TC}},
    \label{eq:lag1-match}
\end{equation}
so that the correlation between adjacent impulse-response coefficients is the same. The upper panels of Fig.~\ref{fig:stable-rbf-vs-tc} show representative impulse responses. Under the TC prior, the decreasing covariance diagonal progressively reduces the typical amplitude of coefficients at large lags. Under the stable RBF prior, no such decrease occurs in the pointwise variance. Instead, most expansion coefficients in \eqref{eq:twoscale-prior} are drawn from increasingly narrow Gaussian components, while rare large-scale activations preserve the constant RBF covariance. Despite this different mechanism, the resulting impulse responses belong to $\ellone$ almost surely.

The lower panels show the distributions of the corresponding normalized frequency-response magnitudes. Each realization is normalized by its maximum magnitude, so that the comparison emphasizes spectral shape rather than overall gain.

%%%%%%%%%%%%%%%%%%%%%

\section{Numerical Experiments}
\label{sec:numexp}

We finally investigate whether the additional flexibility of the stable RBF prior can be useful for impulse-response estimation. We consider FIR approximations with $n_b=200$ coefficients and generate $N=300$ input--output observations using an i.i.d.\ Gaussian input with unit variance and Gaussian measurement noise with variance $\sigma^2=0.5$. For the six systems considered below, this corresponds %to nominal signal-to-noise ratios of approximately $9$--$14$ dB, or equivalently 
to a noise standard deviation about $20$--$35\%$ of that of the noise-free output. The true noise variance is supplied to all estimators. Results are based on $1000$ independent Monte Carlo runs and are reported in terms of
\[
\mathrm{FIT}
=
100\left(
1-\frac{\|\widehat g-g\|_2}{\|g\|_2}
\right).
\]

We compare the stable non-Gaussian RBF prior (sRBF) with Gaussian priors based on the standard RBF and TC covariances. For sRBF, the hyperparameters $(\alpha,\lambda,\beta)$ are estimated using the Monte Carlo expectation-maximization (MCEM) procedure described in \ref{app:srbf-computation}; RBF and TC hyperparameters are estimated by marginal likelihood \citep{RegularizedSI2022}.

For the RBF kernel, the kernel sections are orthonormalized by Gram--Schmidt in their natural temporal order. Since each section belongs to $\ell_1$, the resulting orthonormal basis is $\ell_1$-valued. Moreover, its restriction to the first $n_b$ coordinates gives exactly the lower Cholesky factor of the $n_b\times n_b$ RBF Gram matrix, which is the representation used in the implementation.

\begin{figure*}[t]
\centering
\includegraphics[scale=0.34]{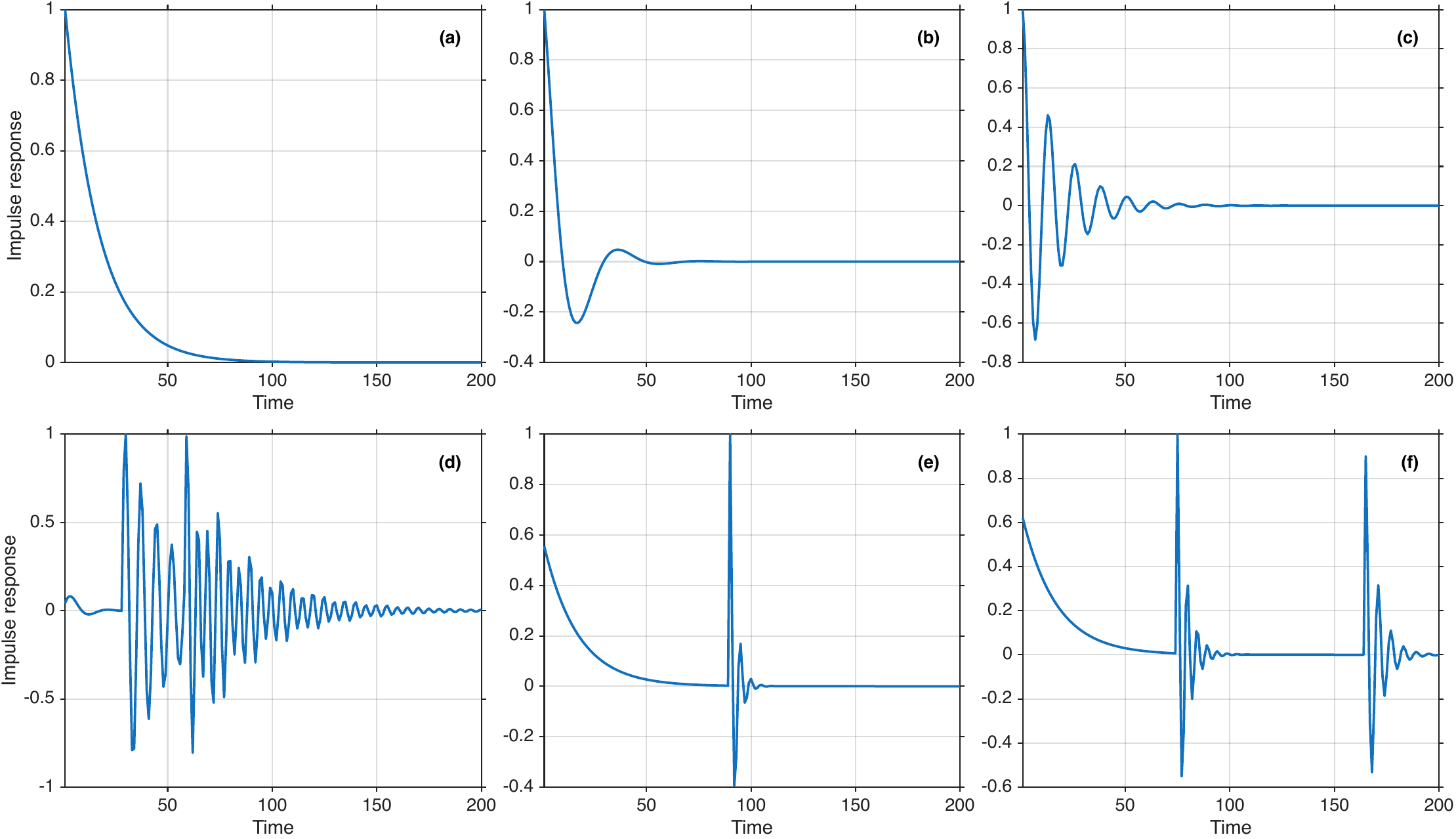}%[width=\linewidth]
\caption{\textbf{Impulse responses used in the Monte Carlo study.}
The six examples progressively include more pronounced oscillatory, delayed, and localized components.}
\label{fig:num-truths}
\end{figure*}

Figure~\ref{fig:num-truths} shows the six impulse responses to be estimated from data. They range from conventional rapidly decaying responses to systems containing oscillatory, localized, or well-separated components at relatively large lags. In particular, the last two examples, although synthetic, are representative
of realistic situations in which relevant dynamics appear at large delays, as in multipath or echo propagation and in delayed resonant or reverberant
responses \citep{ProakisSalehi2008,Kuttruff2016}.

\begin{figure*}[t]
\centering
\includegraphics[scale=0.34]{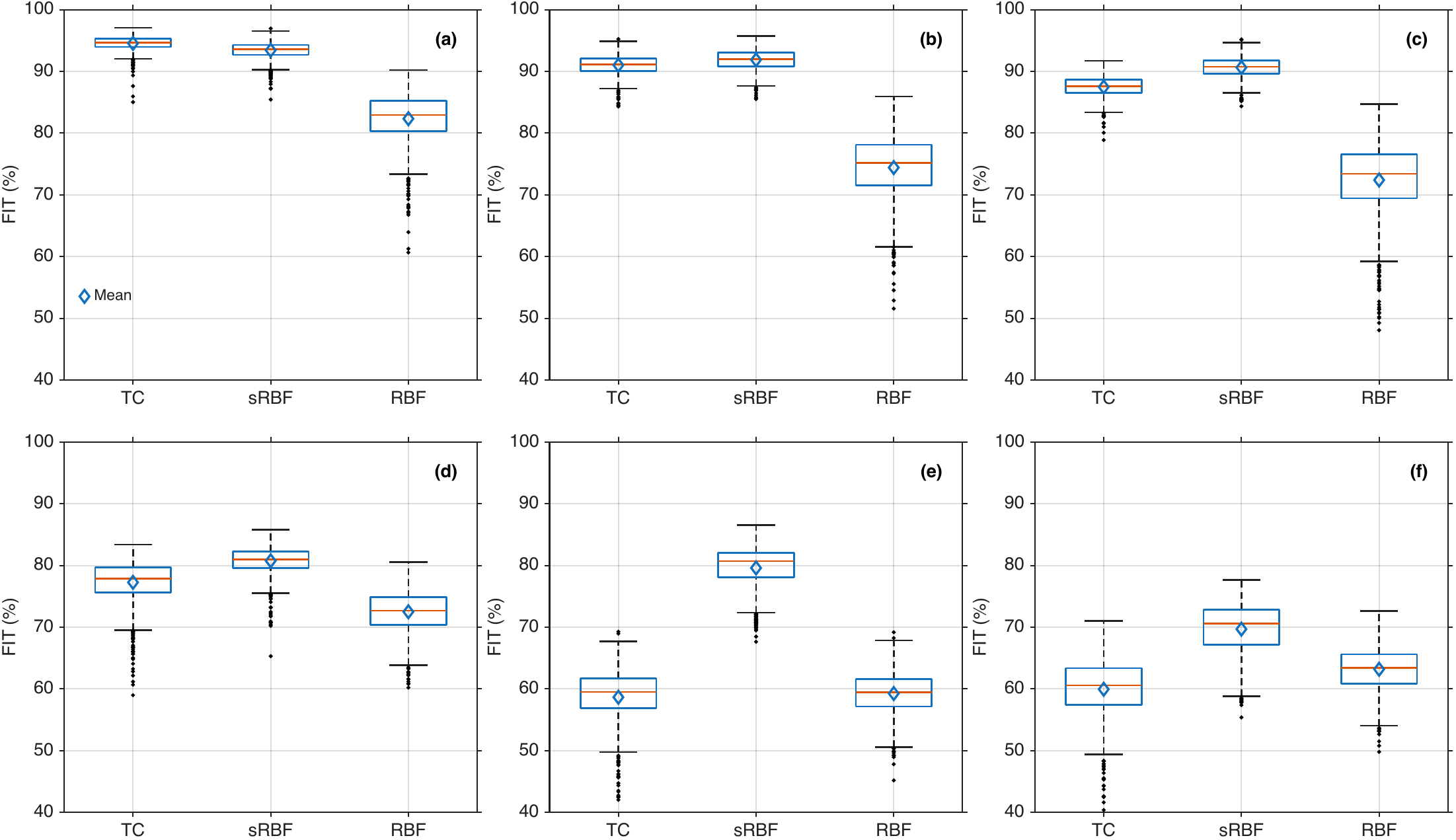}%[width=\linewidth]
\caption{\textbf{Data-driven impulse-response estimation.}
Distribution of FIT over $1000$ Monte Carlo runs for the TC, stable RBF (sRBF), and Gaussian RBF estimators. Hyperparameters are estimated from the observed data by marginal likelihood for TC and RBF and by MCEM for sRBF. Diamonds indicate mean FIT.}
\label{fig:num-datadriven}
\end{figure*}

Figure~\ref{fig:num-datadriven} reports the fully data-driven results. For the more conventional responses in panels (a)--(b), sRBF and TC have comparable performance, with TC slightly better in panel (a). The advantage of sRBF becomes visible in panels (c)--(d), where its mean FIT exceeds that of TC by approximately $3.2$ and $3.5$ points, respectively, and becomes much larger for the late localized structures in panels (e)--(f), with gains of approximately $20.9$ and $9.7$ FIT points. Overall, the Gaussian RBF estimator is the least effective of the three across these examples, as expected since, unlike TC and sRBF, the corresponding Gaussian prior does not encode BIBO stability, although it can outperform TC in individual cases.

The posterior Bernoulli variables provide additional insight into how the two-scale prior adapts to the different impulse responses. Recall from
\eqref{eq:twoscale-coeff} and  \eqref{eq:twoscale-prior} that $a_j$ is the coefficient associated with $\phi_j$, while $B_j=1$ selects its rare, large-scale component and $B_j=0$ its ordinary small-scale component. In panels (c)--(d), the posterior probabilities of such large-scale activations are negligible in almost all Monte Carlo runs. Thus, the good performance of sRBF in these cases does not rely on rare activations.

A markedly different behavior emerges in panels (e)--(f). The following quantities are posterior diagnostics computed separately in each experiment and then averaged over the $1000$ Monte Carlo runs. In panel (e), the posterior expected number of active indicators is on average $2.23$, and the posterior probability of at least two active indicators averages $0.948$. The largest mean inclusion probabilities occur at $j=90$ and $91$, precisely around the delayed ringdown. In panel (f), the posterior expected number of active indicators increases to $4.76$, while $B_{75}$, $B_{76}$, $B_{165}$, and $B_{166}$ receive large mean posterior inclusion probabilities, consistently localizing the two separated delayed components. Hence, in these more challenging examples, the rare large-scale component of the non-Gaussian prior is actively exploited at the temporal locations where additional flexibility is needed.

\begin{figure*}[t]
\centering
\includegraphics[scale=0.34]{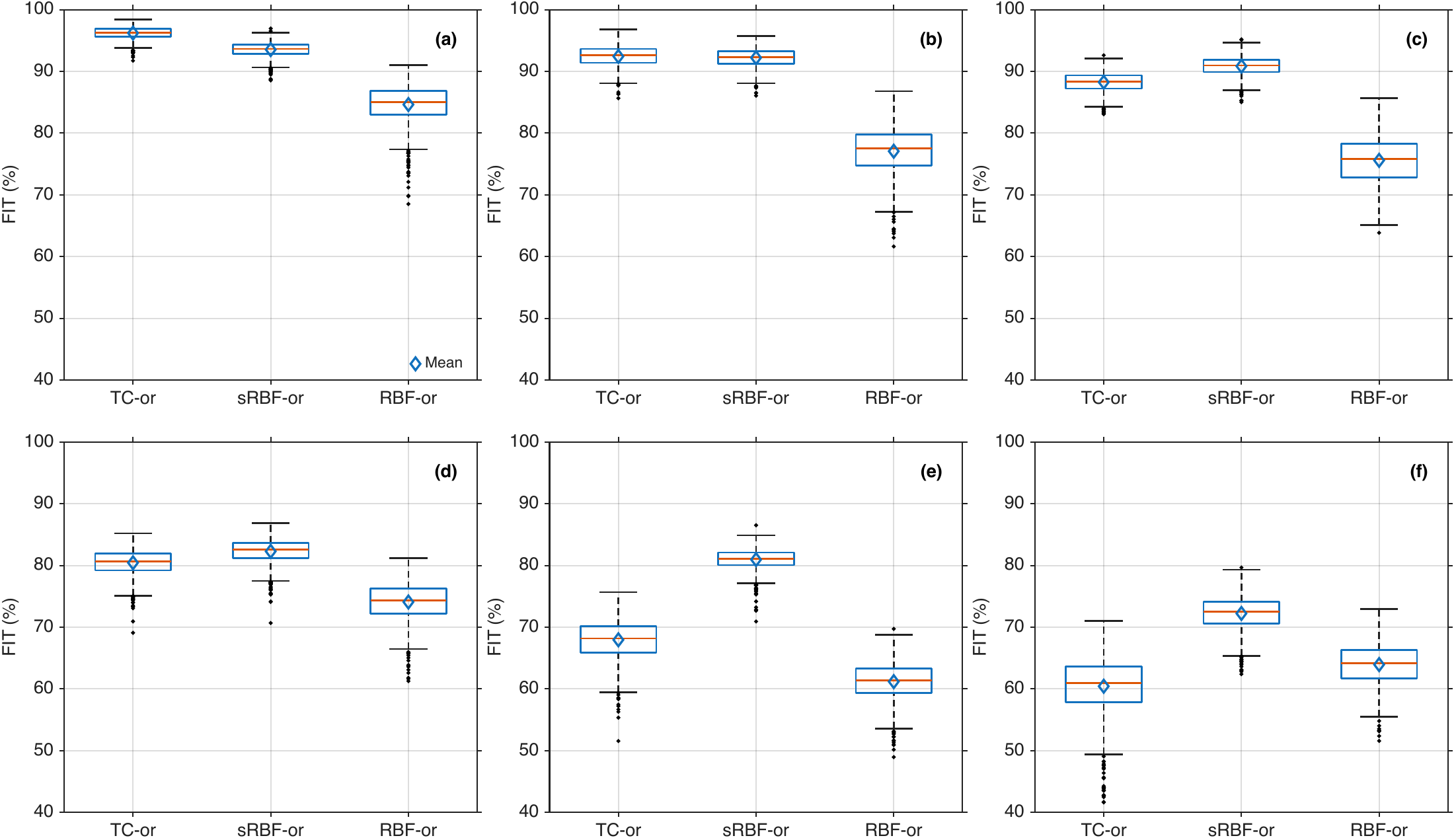}%[width=\linewidth]
\caption{\textbf{Oracle impulse-response estimation.}
Distribution of FIT over $1000$ Monte Carlo runs when the hyperparameters of each method are selected with access to the true impulse response. The main qualitative differences observed in the data-driven comparison persist under oracle tuning.}
\label{fig:num-oracle}
\end{figure*}

Finally, Fig.~\ref{fig:num-oracle} reports oracle results, in which the hyperparameters of each estimator are selected using the true impulse response. The same qualitative picture persists: sRBF remains particularly advantageous for the late and localized responses in panels (e)--(f), while no method uniformly dominates across all six systems. This confirms that the main performance differences are not merely a consequence of hyperparameter estimation, and supports the usefulness of separating the covariance structure from the higher-order mechanism enforcing stability.

\section{Conclusions}
\label{sec:conclusions}

This paper has introduced the notion of a stability-compatible kernel, characterizing the covariance kernels that admit at least one stochastic realization with absolutely summable sample paths. Stability-compatibility is strictly weaker than conventional RKHS stability and therefore enlarges the class of covariance structures that can be used for stable system identification. We have also provided practical sufficient conditions and explicit examples showing that neither Gaussian stability nor RKHS stability is necessary.

This additional freedom can be exploited constructively. We developed non-Gaussian priors that preserve a prescribed covariance while enforcing almost-sure stability, and used this idea to obtain a stable prior based on the stationary RBF kernel. In this way, local dependence and smoothness can be encoded through the covariance, while global stability is enforced through higher-order features of the prior law. The numerical experiments show that this separation can substantially improve impulse-response estimation, especially when relevant dynamics occur at late lags or in separated temporal regions, while remaining competitive on more conventional stable systems.

The price for this additional flexibility is computational. Unlike the Gaussian case, posterior inference is no longer reducible to a linear closed-form estimator, even when the hyperparameters are known, and generally requires more demanding nonlinear Bayesian computations. Our experiments nevertheless show that modern Bayesian inference tools can handle this additional complexity effectively. Thus, moving beyond Gaussianity appears to offer a useful trade-off between computational cost and greater modeling and statistical flexibility in kernel-based system identification.

\appendix

\section{Weighted diagonal Gram bounds}
\label{app:lower-gram}

We prove condition~2 of Proposition~\ref{prop:sections}. We use the
standard RKHS membership criterion of Aronszajn \citep{Aronszajn1950}:
a function $f$ on $\mathbb N$ belongs to $\mathcal H_K$ if and only if
there exists $C\ge0$ such that
\begin{equation}
\left|\sum_{h=1}^N a_h f(n_h)\right|^2
\le
C^2\sum_{i,j=1}^N a_i a_j K(n_i,n_j)
\label{eq:aronszajn-membership}
\end{equation}
for every finite set of distinct indices $n_1,\ldots,n_N$ and every
choice of coefficients $a_1,\ldots,a_N$. The smallest admissible $C$
equals $\|f\|_{\mathcal H_K}$.

Assume that \eqref{eq:weighted-lower-gram} and
\eqref{eq:weighted-sections} hold for some sequence $w_h>0$.
Since the domain is $\mathbb N$, it is enough to verify Aronszajn's
criterion on the leading principal blocks $K_N$: any finite set of
indices is contained in $\{1,\ldots,N\}$ for $N$ sufficiently large,
with the coefficients corresponding to the unused indices set equal
to zero. Hence
\begin{equation}
\sum_{i,j=1}^N a_i a_j K(i,j)
\ge
\sum_{h=1}^N w_h |a_h|^2 ,
\qquad N\ge1.
\label{eq:weighted-leading-bound}
\end{equation}

We first show that every canonical sequence $e_i$, defined by
$e_i(h)=\delta_{ih}$, belongs to $\mathcal H_K$. If $N<i$, then
\[
\sum_{h=1}^N a_h e_i(h)=0.
\]
If $N\ge i$, then
\[
\left|
\sum_{h=1}^N a_h e_i(h)
\right|^2
=
|a_i|^2,
\]
while \eqref{eq:weighted-leading-bound} gives
\[
\sum_{h,j=1}^N a_h a_j K(h,j)
\ge
w_i |a_i|^2.
\]
Thus Aronszajn's criterion yields
\[
e_i\in\mathcal H_K,
\qquad
\|e_i\|_{\mathcal H_K}
\le
\frac{1}{\sqrt{w_i}}.
\]
One thus has $c_{00}\subset\mathcal H_K\cap\ell_1$,
where $c_{00}$ denotes the set of finitely supported sequences.

It remains to show that $c_{00}$ is dense in $\mathcal H_K$.
Fix a kernel section $f=K(\cdot,j)$
and let $f_N$ denote its truncation,
\[
f_N(h)=
\begin{cases}
f(h), & h\le N,\\
0, & h>N.
\end{cases}
\]
Since $f_N\in c_{00}$, it is enough to prove that
$f_N\to f$ in $\mathcal H_K$. Let
\[
c_N
=
\sum_{h=N+1}^{\infty}
\frac{|f(h)|^2}{w_h}.
\]
By \eqref{eq:weighted-sections}, $c_N\longrightarrow0$. 
For every $M>N$ and every $a_1,\ldots,a_M$, Cauchy--Schwarz and
\eqref{eq:weighted-leading-bound} give
\begin{align*}
\left|
\sum_{h=1}^M a_h\bigl(f(h)-f_N(h)\bigr)
\right|^2
&=
\left|
\sum_{h=N+1}^M a_h f(h)
\right|^2\\
&\le
\left(
\sum_{h=N+1}^M w_h|a_h|^2
\right)
\left(
\sum_{h=N+1}^M
\frac{|f(h)|^2}{w_h}
\right)\\
&\le
c_N
\sum_{i,j=1}^M a_i a_j K(i,j).
\end{align*}
For $M\le N$ the left-hand side is zero. Hence Aronszajn's criterion
implies
\[
\|f-f_N\|_{\mathcal H_K}
\le
\sqrt{c_N}
\longrightarrow0.
\]

Thus every kernel section belongs to the
$\mathcal H_K$-closure of $c_{00}$. Since the linear span of the
kernel sections is dense in $\mathcal H_K$, Theorem~\ref{thm:main} then gives stability-compatibility.
%we conclude that
%\[
%\overline{c_{00}}^{\,\mathcal H_K}
%=
%\mathcal H_K.
%\]
%Finally, since
%\[
%c_{00}\subset\mathcal H_K\cap\ell_1,
%\]
%it follows that
%\[
%\overline{\mathcal H_K\cap\ell_1}^{\,\mathcal H_K}
%=
%\mathcal H_K.
%\]
%Theorem~\ref{thm:main} then gives stability-compatibility.

\section{Proof of the stationary spectral criterion}
\label{app:stationary-szego}

\begin{proof}[Proof of Proposition~\ref{prop:stationary-szego-compatible}]
Let $(X_t)_{t\in\mathbb Z}$ be a zero-mean stationary process 
defined on a probability space $(\Omega,\mathcal F,\mathbb P)$, with covariance 
\[
\mathbb E[X_iX_j]=r(i-j)=K(i,j).
\]
As in the proof of Theorem~\ref{thm:main}, the correspondence
\begin{equation}
\mathcal I K(\cdot,n)=X_n,\qquad n\ge1,
\label{eq:stationary-I}
\end{equation}
extends to an isometric isomorphism between $\mathcal H_K$ and
\[
\overline{\operatorname{span}\{X_1,X_2,\ldots\}}^{\,L^2(\Omega)}.
\]

Define the shift operator on the kernel sections by
\[
S K(\cdot,n)=K(\cdot,n+1).
\]
Thus, for every finite linear combination
\[
f=\sum_{n=1}^N a_n K(\cdot,n),
\]
$Sf$ is another sequence 
over $\mathbb N$ defined by
\[
Sf=\sum_{n=1}^N a_n K(\cdot,n+1).
\]
To see that this mapping is well defined and isometric, consider two finite
linear combinations
\[
f=\sum_{i=1}^N a_iK(\cdot,i),
\qquad
g=\sum_{j=1}^M b_jK(\cdot,j).
\]
By stationarity,
\[
\begin{aligned}
\langle Sf,Sg\rangle_{\mathcal H_K}
&=
\sum_{i=1}^N\sum_{j=1}^M
a_i b_j K(i+1,j+1)\\
&=
\sum_{i=1}^N\sum_{j=1}^M
a_i b_j K(i,j)\\
&=
\langle f,g\rangle_{\mathcal H_K}.
\end{aligned}
\]
Hence $S$ is well defined on the span of the kernel sections and preserves
inner products there. Since this span is dense in $\mathcal H_K$, $S$
extends uniquely by continuity to an isometry
\[
S:\mathcal H_K\longrightarrow\mathcal H_K.
\]
Its range is
\[
S\mathcal H_K
=
\overline{\operatorname{span}
\{K(\cdot,2),K(\cdot,3),\ldots\}}^{\,\mathcal H_K}.
\]
More generally,
\begin{equation}
S^j\mathcal H_K
=
\overline{\operatorname{span}
\{K(\cdot,n):n\ge j+1\}}^{\,\mathcal H_K},
\qquad j\ge0.
\label{eq:shifted-rkhs}
\end{equation}
%Notice that both $\mathcal H_K$ and $S\mathcal H_K$ consist of sequences
%defined on the same domain $\mathbb N$. The operator $S$ shifts the index
%of the kernel sections, rather than the domain of the sequences themselves.
In general, $S\mathcal H_K$ may be a proper subspace of $\mathcal H_K$.

Now define the orthogonal residual
\[
e_1
:=
K(\cdot,1)-P_{S\mathcal H_K}K(\cdot,1).
\]
Because $\mathcal I$ is an isometry,
\[
\mathcal I e_1
=
X_1-
P_{\overline{\operatorname{span}\{X_2,X_3,\ldots\}}}X_1
=:\varepsilon_1.
\]
Thus $\varepsilon_1$ is the one-step innovation of the time-reversed
stationary process, and
\begin{equation}
\|e_1\|_{\mathcal H_K}^2
=
\operatorname{Var}(\varepsilon_1).
\label{eq:innovation-variance-isometry}
\end{equation}
By the Szeg\H{o}--Kolmogorov prediction formula \citep{Bingham2012},
\[
\operatorname{Var}(\varepsilon_1)
=
\exp\!\left\{
\int_{-\pi}^{\pi}
\log f(\omega)\,\frac{d\omega}{2\pi}
\right\},
\]
so the Szeg\H{o} condition is equivalent to
\begin{equation}
e_1\neq0.
\label{eq:e1-nonzero}
\end{equation}

Consider now the RKHS tail
\begin{equation}
\mathcal T
:=
\bigcap_{N\ge1}
\overline{\operatorname{span}
\{K(\cdot,n):n\ge N\}}^{\,\mathcal H_K}.
\label{eq:T-rkhs}
\end{equation}
Under the isometry \eqref{eq:stationary-I}, $\mathcal T$ corresponds,
after time reversal, to the classical remote-past subspace
\citep{LindquistPicci2015}. Since the spectral measure is absolutely continuous and the
Szeg\H{o} condition holds, the stationary process is purely
nondeterministic. Hence its remote-past subspace is trivial
\citep{Bingham2012}. Through the isometry \eqref{eq:stationary-I}, this is
precisely the condition $\mathcal T=\{0\}$.
Thus the two spectral assumptions correspond exactly to
\[
e_1\neq0,
\qquad
\mathcal T=\{0\}.
\]

Set $\mathcal W:=\mathcal H_K\ominus S\mathcal H_K$.
Since
\[
\mathcal H_K
=
\overline{
\operatorname{span}\{K(\cdot,1)\}+S\mathcal H_K
}^{\,\mathcal H_K},
\]
and $e_1$ is the orthogonal residual of $K(\cdot,1)$ with respect to
$S\mathcal H_K$, while $e_1\neq0$, it follows that
\[
\mathcal W=\operatorname{span}\{e_1\}.
\]

The Wold decomposition theorem for the isometry $S$ yields
\[
\mathcal H_K
=
\overline{\operatorname{span}
\{S^j\mathcal W:\,j\ge0\}}^{\,\mathcal H_K}
\oplus
\left(
\bigcap_{j\ge0}S^j\mathcal H_K
\right).
\]
By \eqref{eq:shifted-rkhs} and \eqref{eq:T-rkhs},
\[
\bigcap_{j\ge0}S^j\mathcal H_K
=
\mathcal T.
\]
Since $\mathcal T=\{0\}$ and
$\mathcal W=\operatorname{span}\{e_1\}$, we therefore obtain
\begin{equation}
\mathcal H_K
=
\overline{\operatorname{span}
\{e_1,Se_1,S^2e_1,\ldots\}}^{\,\mathcal H_K}.
\label{eq:innovation-dense-rkhs}
\end{equation}

It remains to show that these vectors have finite support, i.e. $e_i \in c_{00}$ for any $i$. Since
\[
e_1\perp S\mathcal H_K
\]
and $K(\cdot,n)\in S\mathcal H_K$ for every $n\ge2$, the reproducing
property gives
\[
e_1(n)
=
\langle e_1,K(\cdot,n)\rangle_{\mathcal H_K}
=
0,
\qquad n\ge2.
\]
More generally, let
\[
e_j:=S^{j-1}e_1,
\qquad j\ge1.
\]
For $n>j$, we have
\[
K(\cdot,n)
=
S^{j-1}K(\cdot,n-j+1).
\]
Hence, since $S$ is an isometry,
\[
\begin{aligned}
e_j(n)
&=
\left\langle
S^{j-1}e_1,K(\cdot,n)
\right\rangle_{\mathcal H_K}\\
&=
\left\langle
S^{j-1}e_1,
S^{j-1}K(\cdot,n-j+1)
\right\rangle_{\mathcal H_K}\\
&=
\langle
e_1,K(\cdot,n-j+1)
\rangle_{\mathcal H_K}\\
&=
0,
\end{aligned}
\]
because $n-j+1\ge2$. Therefore
\[
\operatorname{supp}(e_j)\subseteq\{1,\ldots,j\},
\qquad j\ge1,
\]
and hence every $e_j$ belongs to $\ell_1$. This, together with
 \eqref{eq:innovation-dense-rkhs}, completes the proof.
\end{proof}

\section{Bayesian computation for the stable RBF prior}
\label{app:srbf-computation}

We briefly describe the computational scheme used for the stable RBF
estimator. Consider an FIR truncation of length $n_b$ and write the
input--output model as
\begin{equation}
    y = U g + e,
    \qquad
    e\sim\mathcal N(0,\sigma^2 I).
    \label{eq:app-data-model}
\end{equation}
For a given RBF shape parameter $\alpha$, let
\[
    K_\alpha=\Phi_\alpha\Phi_\alpha^\top
\]
be the Cholesky factorization of the $n_b\times n_b$ RBF Gram matrix.
The truncated stable RBF prior can then be written as
\begin{equation}
    g=\sqrt{\lambda}\,\Phi_\alpha a,
    \qquad
    a\mid B\sim\mathcal N(0,D_B),
    \label{eq:app-srbf-finite}
\end{equation}
where $B=(B_1,\ldots,B_{n_b})$, $B_j\sim\mathrm{Bernoulli}(q_j)$,
and
\[
    D_B=\operatorname{diag}(s_1^2,\ldots,s_{n_b}^2),
    \qquad
    s_j^2=
    \begin{cases}
        c_j^2, & B_j=0,\\
        d_j^2, & B_j=1.
    \end{cases}
\]
The quantities $q_j,c_j,d_j$ are defined as in
Section~\ref{sec:twoscale}. In the numerical experiments $q_j$ is fixed,
so that the unknown hyperparameters are
\[
    \theta=(\alpha,\lambda,\beta).
\]

A useful feature of this representation is that the Gaussian coefficients
$a$ can be integrated out analytically. Conditional on $B$,
\begin{equation}
    y\mid B,\theta
    \sim
    \mathcal N(0,\Sigma_B(\theta)),
    \qquad
    \Sigma_B(\theta)
    =
    \sigma^2 I+
    \lambda U\Phi_\alpha D_B\Phi_\alpha^\top U^\top .
    \label{eq:app-collapsed-cov}
\end{equation}
We therefore sample only the Bernoulli indicators using a collapsed Gibbs
scheme \citep{GelfandSmith1990,LiuWongKong1994}, i.e., 
after analytically integrating out the Gaussian coefficients $a$, inference
on the latent variables is carried out through a 
sampler acting only on the Bernoulli indicators $B$.
Specifically, each
$B_j$ is updated from
\begin{equation} \small
\mathbb P(B_j=b\mid B_{-j},y,\theta)
\propto
q_j^b(1-q_j)^{1-b}
|\Sigma_{B_j=b}|^{-1/2}
\exp\!\left(
-\frac12 y^\top\Sigma_{B_j=b}^{-1}y
\right),
\label{eq:app-gibbs}
\end{equation}
with $b\in\{0,1\}$.
Changing a single indicator modifies $\Sigma_B$ by a rank-one term.
Hence the two probabilities in \eqref{eq:app-gibbs} can be evaluated
efficiently using rank-one determinant and inverse updates, rather than
factorizing a new covariance matrix for every Gibbs update.

For each retained configuration $B$, the conditional posterior mean of the
impulse response is available in closed form:
\begin{equation}
m_g(B;\theta)
=
\lambda\Phi_\alpha D_B\Phi_\alpha^\top U^\top
\Sigma_B(\theta)^{-1}y .
\label{eq:app-condmean}
\end{equation}
The stable RBF estimate is obtained by Rao--Blackwellization,
\begin{equation}
    \widehat g
    =
    \mathbb E[g\mid y,\theta]
    \simeq
    \frac{1}{M}\sum_{m=1}^{M}
    m_g(B^{(m)};\theta),
    \label{eq:app-rbmean}
\end{equation}
which avoids explicitly sampling the Gaussian coefficients $a$ and
reduces Monte Carlo variability \citep{LiuWongKong1994}.

When the hyperparameters are unknown, we estimate them by a Monte Carlo
EM scheme \citep{DempsterLairdRubin1977,WeiTanner1990}. At iteration
$r$, the E-step generates posterior samples
$B^{(1)},\ldots,B^{(M)}$ using the collapsed Gibbs sampler at
$\theta^{(r)}$. The M-step then maximizes
\begin{equation}
    Q_M(\theta\mid\theta^{(r)})
    =
    -\frac{1}{2M}
    \sum_{m=1}^{M}
    \left[
       \log|\Sigma_{B^{(m)}}(\theta)|
       +
       y^\top\Sigma_{B^{(m)}}(\theta)^{-1}y
    \right],
    \label{eq:app-mcem}
\end{equation}
up to terms independent of $\theta$. The maximization is performed
continuously over $(\alpha,\lambda,\beta)$ using multiple starting points.
After convergence, a longer collapsed Gibbs run at the estimated
hyperparameters is used in \eqref{eq:app-rbmean} to compute the final
posterior-mean estimate. When the hyperparameters are known, the MCEM step
is simply omitted.

\end{document}